\documentclass[letterpaper, 10 pt, conference]{ieeeconf}

\IEEEoverridecommandlockouts
\usepackage{color}
\usepackage{multirow}
\usepackage{booktabs}
\usepackage{url}
\usepackage{subfigure}
\usepackage{threeparttable}
\usepackage{cite}
\makeatletter
\let\NAT@parse\undefined
\makeatother
\usepackage{hyperref}[colorlinks,linkcolor=blue]
\usepackage[ruled,linesnumbered]{algorithm2e}
\renewcommand{\baselinestretch}{0.966}
\usepackage[margin=0.75in]{geometry}
\usepackage{array}
\newcolumntype{L}[1]{>{\raggedright\arraybackslash}p{#1}}
\usepackage{hyperref}[colorlinks,linkcolor=blue]
\usepackage[ruled,linesnumbered]{algorithm2e}
 \renewcommand{\baselinestretch}{0.966}

\usepackage{booktabs}
\usepackage{enumerate}
\usepackage{amssymb}
\usepackage{leftidx}
\usepackage{amsfonts}
\usepackage{amsmath}
\usepackage{bm}
\usepackage{booktabs}
\usepackage{setspace}
\usepackage{makecell}
\usepackage{setspace}
\usepackage{siunitx}
\usepackage{float}
\usepackage{subfigure}
\usepackage[pdftex]{graphicx}
\usepackage{cite}
\usepackage{siunitx}
\usepackage{cleveref}

\graphicspath{{figs/}}
\usepackage{graphicx}
\graphicspath{{figs/}}

\usepackage{amsthm}
\newtheorem{theorem}{Theorem}

\newcommand{\etal}{\textit{et al}.}

\DeclareSIUnit{\mps}{m/s} 

\begin{document}

\title{\LARGE \bf
Accelerating Large-scale Bundle Adjustment for LiDAR Mapping \\via Parallel Computing
}

\author{Yixi Cai$^{1}$, Rundong Li$^2$, Yuhan Xie$^3$, Qingwen Zhang$^1$, Patric Jensfelt$^1$, and Fu Zhang$^2$
 \thanks{
 $^1$Y. Cai, Q. Zhang, and P. Jensfelt are with the Division of Robotics and Perception and Learning, KTH Royal Institute of Technology, Stockholm, Sweden.
 \texttt{\{yixica, qingwen, patric\}@kth.se}.
 $^2$R. Li and F. Zhang are with the Department of Mechanical Engineering, the University of Hong Kong. 
 $^3$Yuhan Xie is with the School of Computing and Data Science, the University of Hong Kong
 \texttt{\{rdli10010,yuhanxie\}@connect.hku.hk}, \texttt{fuzhang@hku.hk} 
 }
 \thanks{Corresponding author: Fu Zhang}
}

\maketitle
\pagestyle{empty} 
\thispagestyle{empty} 

\begin{abstract}
LiDAR bundle adjustment is widely utilized in mapping to construct globally consistent point cloud maps. In this paper, we propose the first fully parallel computing framework to accelerate LiDAR bundle adjustment for large-scale mapping, incorporating three key techniques. First, we design an adaptive, asynchronous data loading strategy to efficiently process large-scale point cloud datasets on memory-constrained GPUs. Secondly, we present a novel bottom-up voxelization method for extracting planar features, enabling fully parallelized pre-processing. Thirdly, we build upon a majorization-minimization formulation to accelerate compute-intensive tasks in the optimization via parallel computation, including the computation of residuals, Jacobian and Hessian matrices, and a parallel increment solver. To support our design, we provide both theoretical and experimental analysis of the time complexity of our approach. Extensive benchmarking on large-scale public datasets across various computational platforms validates the robustness and adaptability of our approach, achieving up to a tenfold improvement in computational efficiency while preserving mapping accuracy comparable to state-of-the-art methods. To benefit future research, the implementation code is available on GitHub\footnote{https://github.com/Ecstasy-EC/GPU-BALM}. 
\end{abstract}

\section{Introduction}
\label{sec:intro}

LiDAR (Light Detection and Ranging) sensors are a critical sensing technology, providing high-resolution, accurate geometric data of environments with an extended detection range. They are widely used in creating detailed 3D digital twins for applications such as augmented reality (AR) and virtual reality (VR)\cite{cipresso2018past}, as well as in autonomous driving~\cite{levinson2011towards}, robotics\cite{ren2025safety}, and large-scale urban mapping~\cite{lim2024kiss}, where precise 3D representations are vital for navigation and planning.

LiDAR bundle adjustment (BA) is essential for building globally consistent 3D representations~\cite{liu2021balm} by jointly optimizing sensor poses and point clouds, minimizing drift, and enhancing alignment across multiple scans. Unlike pairwise registration, BA processes all scans simultaneously, which directly influences the accuracy of environmental reconstructions.

As the scale of LiDAR datasets has expanded from small campus-scale datasets~\cite{ramezani2020newer} to large city-scale datasets~\cite{jung2024helipr}, performing bundle adjustment on such large-scale data presents significant computational challenges. Traditional approaches predominantly rely on sequential computation schemes\cite{liu2021balm}, where the pre-processing of LiDAR point clouds is computationally expensive, with its processing time increasing linearly with the number of scans and the density of LiDAR sensors, which often reaches a total of billions of points for a sequence. Furthermore, the optimization process itself is highly demanding, as it typically involves iterative nonlinear solvers whose computational cost scales poorly with the growing number of scans and points~\cite{liu2023efficient}. Although some methods employ multi-threading\cite{liu2023large} or decentralized CPU processing to achieve moderate parallelism~\cite{li2026efficient}, these methods remain insufficient for handling large-scale datasets efficiently, limiting their applicability in large-scale mapping scenarios.

Given the rapid advancements in high-performance computing~\cite{epoch2022trends}, GPU-based parallel processing offers a promising approach for accelerating LiDAR bundle adjustment. However, acceleration cannot be readily achieved through a straightforward migration of algorithms from CPU to GPU. Instead, a carefully crafted design is required, taking into account memory allocation, data flow, and problem formulation to fully harness the computational power of GPUs. This work presents the first fully parallel LiDAR bundle adjustment framework, designed to improve computational efficiency and scalability compared to conventional methods. Our approach addresses key challenges in parallel computation through three primary innovations. First, we introduce an adaptive asynchronous loading strategy to mitigate the constraints of limited graphical memory in modern GPUs when handling large-scale point clouds, often on the order of tens of gigabytes. Second, in contrast to existing methods that predominantly employ a top-down voxelization architecture for point cloud processing \cite{liu2021balm, liu2023efficient, li2026efficient}, we propose a novel bottom-up voxelization approach to extract planar features. This method enables fully parallelized pre-processing, enhancing computational efficiency. Third, we develop a fully parallelized optimization process based on the majorization-minimization formulation introduced in BALM3 \cite{li2026efficient}. By integrating these techniques, our framework operates entirely on a single GPU, eliminating the need for CPU-GPU communication once data is transferred, and offers potential scalability to multi-GPU configurations. The contributions of this work are summarized as follows:

\begin{itemize}
 \item We propose an adaptive, asynchronous data loading strategy that enables efficient processing of large-scale point cloud datasets on GPUs with constrained memory capacities, addressing the challenges posed by datasets exceeding typical GPU memory limits.
 \item We develop a novel bottom-up voxelization method for extracting planar features from point clouds, a computationally intensive component of LiDAR processing pipelines, enabling fully parallelized pre-processing to enhance efficiency.
 \item We design fully parallelized computational algorithms for the optimization process, leveraging a majorization-minimization formulation, thereby significantly reducing computational complexity.
 \item We present a theoretical and experimental analysis of the time complexity of our proposed parallel computing framework, providing formal justification for its design.
 \item We conduct extensive benchmarking on large-scale public datasets across various computational platforms, achieving up to a tenfold improvement in computational efficiency while maintaining accuracy comparable to state-of-the-art methods. Our implementation code will be released on GitHub upon acceptance.
\end{itemize}

\section{Related Works}
\label{sec:related}
\subsection{LiDAR Bundle Adjustment}
Recent advances in multi-view LiDAR registration have seen growing interest in point cloud bundle adjustment. Early works like Kaess \etal~\cite{kaess2015simultaneous, hsiao2017keyframe} and Ferrer \textit{et al.}~\cite{ferrer2019eigen} pioneered plane-feature-based BA by minimizing point-to-plane distances. Kaess \textit{et al.} jointly optimized scan poses and plane parameters via Levenberg–Marquardt (LM) optimization, while Ferrer \textit{et al.} eliminated plane parameters by reformulating the problem into an eigenvalue minimization task solved with a first-order optimizer. As a result, the method is termed ``Eigen-factor''.

Later, Liu \textit{et al.} introduced the BALM series\cite{liu2021balm, liu2023efficient}. In its initial version\cite{liu2021balm}, BALM minimizes the point-to-plane distance and analytically solves the plane parameters using closed-form solutions before the numerical optimization similar to Eigen-factor. Moreover, they utilized a second-order solver with analytical Jacobian and Hessian matrices which achieved faster convergence. Further refinement led to BALM2 \cite{liu2023efficient}, which introduced point clusters to avoid enumerating raw points during cost function and derivative computations. By deriving cost, Jacobian, and Hessian terms directly from point clusters rather than individual points, BALM2 significantly improved computational efficiency without loss of accuracy.

Following BALM2, Li \textit{et al.} recently introduced BALM3\cite{li2026efficient}. BALM3 leverages the majorization-minimization method to decouple LiDAR poses during the LM optimization process. By doing so, the time complexity of the optimization is reduced from cubic to linear, significantly enhancing efficiency, particularly for large-scale datasets. Moreover, the decoupling of LiDAR poses enables parallel computation across key stages of the pipeline, including feature extraction, cost and derivatives evaluation, and Hessian matrix solving. This inherent parallelizability makes BALM3 highly suitable for further acceleration through parallel computing, providing the foundation for our paper.

\subsection{Parallel Computing for LiDAR Mapping}
The potential of parallel computing has been harnessed in recent years, yielding substantial acceleration in LiDAR mapping. Early applications of parallel computing techniques in LiDAR mapping often accelerated only specific components within their frameworks, while the majority of computations relied on sequential processing. For instance, SuMA~\cite{behley2018efficient} enhances correspondence matching by rendering surfels\cite{pfister2000surfels} in parallel on the GPU. Elastic-LiDAR Fusion~\cite{park2018elastic} and Droeschel \textit{et al.}\cite{droeschel2018efficient} primarily leverage the GPU for scene reconstruction rather than core mapping tasks. Chen \textit{et al.}\cite{chen2021gpu} utilize the GPU to accelerate wavefront updates in truncated signed distance fields (TSDF), while retaining ray casting and occupancy updates on the CPU. In PGO-LIOM~\cite{shen2022pgo}, Shen \textit{et al.} employ a sampling-based, gradient-free optimization method to address the nonlinear cost function, accelerated via parallel computing. Although these approaches exploit parallel architectures to enhance specific aspects of their frameworks, their overall efficiency remains constrained by sequential computation on CPUs.

Subsequently, researchers have focused on shifting greater computational workloads to GPUs to achieve enhanced acceleration. Chen \textit{et al.} extend their earlier work~\cite{chen2021gpu} into an incremental Euclidean distance transform framework, where data is entirely stored in GPU memory and processed fully through parallel computation~\cite{chen2022gpu}. Alexander \textit{et al.} introduce nvblox~\cite{millane2024nvblox}, building upon voxblox~\cite{oleynikova2017voxblox}, by implementing all computations on the GPU in parallel to enable real-time motion planning. Modern GPU-based computation techniques are leveraged in~\cite{koide2021globally}, which parallelly minimizes the matching cost factors of General-ICP (G-ICP)\cite{segal2009generalized} in a pose graph optimization framework. 

This work draws inspiration from MegBA~\cite{ren2022megba}, which accelerates visual bundle adjustment through parallel computing on GPUs. Our framework targets LiDAR bundle adjustment for point clouds, a process that imposes a significant computational burden due to the large volume of points, up to billions, required to ensure global consistency. By leveraging a decoupled formulation based on the majorization-minimization method~\cite{li2026efficient}, all computations within this framework are executed on the GPU, with the computational pipeline carefully designed using modern GPU techniques to fully exploit the processing capabilities of GPUs.

\section{LiDAR Bundle Adjustment via Majorization-Minimization }
\label{sec:formulation}
Building on the BALM framework\cite{liu2021balm, liu2023efficient}, we formulate the point-to-plane residual as the cost function using point clusters\cite{liu2023efficient}. Assuming there are $N_\mathtt{f}$ plane features being observed by $N_\mathtt{T}$ LiDAR scans with poses $\mathtt T_j=(\mathbf R_j, \mathbf t_j)\in SE(3)$, $(j=1,...,N_\mathtt{T})$, the cost function is
\begin{align}
    \min_{\mathtt T_j \in SE(3), \forall j} c(\mathbf T) &= \sum_{i=1}^{N_\mathtt{f}} \underbrace{ \lambda_{\min} \left( \frac{1}{N_i} \mathbf P_i - \frac{1}{N_i^2} \mathbf v_i \mathbf v_i^T \right)}_{c_i (\mathbf T)}, \; \nonumber \\
    \quad \mathbf P_i &= \sum_{j=1}^{N_\mathtt{T}} \mathbf P_{ij}, \mathbf v_i = \sum_{j=1}^{N_\mathtt{T}} \mathbf v_{ij}  \label{BA-formulation-raw},
\end{align}
where $\lambda_{\text{min}} ( \cdot )$ denotes the minimal eigenvalue of the input matrix, $N_i$ is the total number of raw points on the $i$-th plane feature observed by $N_\mathtt{T}$ LiDAR scans, and for each scan $j$
\begin{align}
    \label{eq:cluster}
        {\mathbf P}_{ij} &= \mathbf R_j \mathbf P_{f_{ij}} \mathbf R_j^T + \mathbf R_j \mathbf v_{f_{ij}} \mathbf t_j^T + \mathbf t_j \mathbf v_{f_{ij}}^T \mathbf R_j^T + N_{ij} \mathbf t_j \mathbf t_j^T, \; \nonumber \\
    \quad {\mathbf v}_{ij} &= \mathbf R_j \mathbf v_{f_{ij}} + N_{ij} \mathbf t_j \; \nonumber \\
    \mathbf P_{f_{ij}} &= \sum_{k=1}^{N_{ij}} \mathbf p_{f_{ijk}} \mathbf p_{f_{ijk}}^T, \quad
    \mathbf v_{f_{ij}} = \sum_{k=1}^{N_{ij}} \mathbf p_{f_{ijk}}
\end{align}
with $N_{ij}$ being the total number of raw points on the $i$-the planar feature observed by the $j$-th scan, and $\mathbf p_{f_{ijk}}$ being the $k$-th raw point of them. The $i$-th planar feature is represented by the tuple $(P_{f_{ij}}, v_{f_{ij}}, N_{ij})$, which we denote as $i$-th ``point cluster''.

To enable parallel computation, we employ BALM3's surrogate function\cite{li2026efficient} for indirect optimization:
\footnotesize
\begin{align}\label{eq:surrogate}
    c_{M_i} (\mathbf T | \mathbf T^{(l)}) \triangleq& \sum_{j=1}^{N_T} \left( \frac{1}{N_i} {\mathbf u_{\min}^{(l)}}^T {\mathbf P}_{ij} \mathbf u_{\min}^{(l)} - \frac{4 z^{(l)}}{N_i^2} {\mathbf u_{\min}^{(l)}}^T {\mathbf v}_{ij}  \right) \; \nonumber \\
    &+ \frac{4}{N_i^2} {z^{(l)}}^2
\end{align}
\normalsize
where $\mathbf u_{\min}^{(l)}$ is the eigenvector associated to the minimum eigenvalue of the matrix $\frac{1}{N_i} \mathbf P_i^{(l)} - \frac{1}{N_i^2} \mathbf v_i^{(l)} {\mathbf v_i^{(l)}}^T$ and $z^{(l)} = \frac{1}{2}\sum_{j=1}^{M_p}{\mathbf u_{\min}^{(l)}}^T {\mathbf v}_{ij}^{(l)}$ with $\mathbf P_i^{(l)}, \mathbf v_i^{(l)}, {\mathbf v}_{ij}^{(l)}$ evaluated at $\mathbf T^{(l)}$, respectively. $l$ denotes $l$-th iteration of the optimization.
For the corresponding first-order and second-order derivatives, we refer the reader to \cite{li2026efficient}.
\section{Parallel Computation}
\begin{figure*}
    \centering
    \setlength\abovecaptionskip{-0.5\baselineskip}    
    \includegraphics[width=\linewidth]{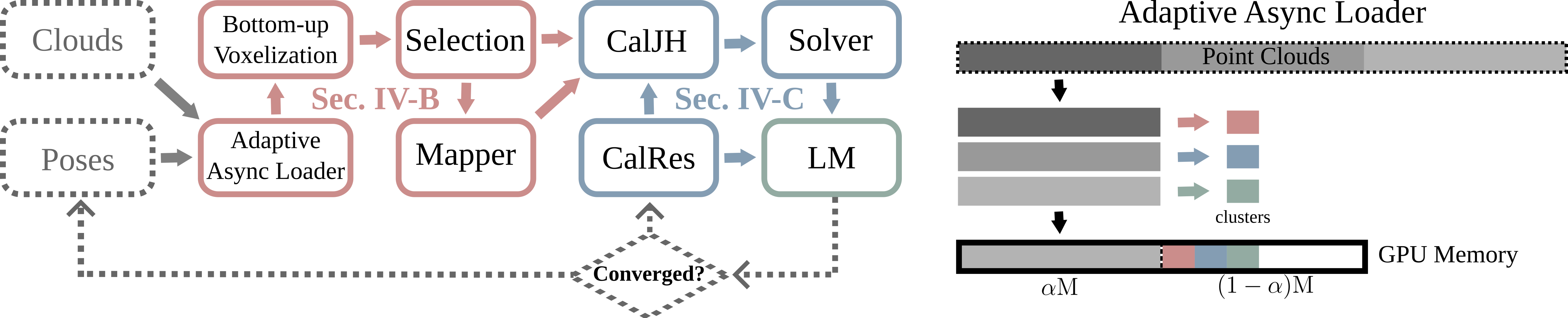}
    \caption{Left: Parallel Computation Flow. Boxes with dotted lines indicate sequential computations performed on the CPU, whereas those with solid lines denote parallel computations executed on the GPU. The red boxes represent the preprocessing process, as described in \Cref{subsec:preprocess}, while the blue boxes indicate the optimization process, as detailed in Section~\ref{subsec:opt}. The green box with dotted lines denotes the Levenberg–Marquardt (LM) algorithm serving as the optimizer. Right: Illustration of Adaptive Asynchronous Data Loader}
    \label{fig:flow}
    \vspace{-0.4cm}
\end{figure*}
This section presents our parallel computing framework, as illustrated in \Cref{fig:flow}. We begin by briefly introducing the fundamental parallel computing primitives frequently utilized within our framework. We then detail our parallel design for preprocessing and optimization, both of which are executed fully in parallel on the GPU.

\subsection{Standard Primitives}
\label{subsec:primitives}
To achieve a fully parallel processing framework, our method extensively utilizes four standard parallel computing primitives: \texttt{sort\_by\_key}, \texttt{reduce\_by\_key}, \texttt{exclusive\_scan}, and \texttt{scatter}. These primitives are fundamental for efficiently processing large-scale data in parallel.  

\begin{itemize}
    \item [1)]\texttt{sort\_by\_key}: This primitive takes an array of keys and corresponding values as input and sorts the key-value pairs in parallel based on the keys. A common implementation employs parallel radix sort~\cite{lee2002partitioned}.
    \item [2)]\texttt{reduce} \& \texttt{reduce\_by\_key}: The term ``reduce'' refers to parallel summation in the field of parallel computing. The primitive \texttt{reduce\_by\_key} performs associative binary operations, such as summation, minimum, or maximum, on consecutive values sharing the same key. The results are then stored in a compact output array, whose length corresponds to the number of unique keys. 
    \item [3)]\texttt{exclusive\_scan}: This primitive computes the prefix sum of an input array, where each element in the output represents the total sum of all preceding elements, excluding the current term.
    \item [4)]\texttt{scatter}: This primitive is used to re-arrange an array $\mathtt{R_A}$ based on a given mapping array $\mathtt{R_B}$.
\end{itemize}

All these primitives are efficiently implemented in the Thrust library~\cite{bell2012thrust}, removing the need for custom re-implementation and enabling high-performance execution on modern GPU architectures. For parallel computations that cannot be supported by standard primitives, we design customized parallel kernels, denoted as ``Parallel For'' in our algorithm.
\subsection{Preprocessing}
\label{subsec:preprocess}
As illustrated in \Cref{fig:flow}, the preprocessing (denoted by red boxes) commences with the adaptive asynchronous data loader to transfer point cloud data from CPU to GPU. Following that, we index the point clouds via a bottom-up voxelization module to enable parallel planar feature extraction.
We define a multi-resolution hierarchy with $\mathtt{L}$ levels.
For the voxel at the $k$-th level ($k$$=$$1$, ..., $\mathtt{L}$), the resolution is obtained by $\mathtt d^k = 2^{k-1}\mathtt d$, where $\mathtt d$ is the minimal resolution.
Subsequently, points within voxels exhibiting robust good features are selected. Finally, two index mappers are prepared to facilitate parallel computation in the optimization process.
%
\subsubsection{Adaptive Asynchronous Data Loader}
In the processing of large-scale point cloud datasets, the substantial size of point clouds often exceeds GPU memory capacity. Even if the point clouds can be stuffed into the GPU memory, they consume the memory needed for other subsequent processing. To mitigate the limited memory constraints, we take advantage of the compact cluster representation in \Cref{eq:cluster}. Specifically, we partition the GPU memory into two segments using a ratio $\alpha$, as illustrated in the right panel of \Cref{fig:flow}: one segment is allocated for loading point clouds, while the other serves as a buffer for cluster storage and processing. Point clouds stored in CPU memory are segmented into batches and transferred to the first GPU memory segment. Subsequently, we merge points observed at the same pose within the same voxel at the minimum resolution into clusters, as introduced below. 

Each point $p_i = \{x_i, y_i, z_i\}$ from the raw point cloud $\mathtt{P}$ is transformed into the world coordinate as $p^w_i=\{x_i^w, y_i^w, z_i^w\}$ and indexed by the minimum resolution $\mathtt{d}$ using the following equation: 
\begin{equation}
    \mathtt{I}_x^1[i] =\lfloor x_i^w/\mathtt{d}\rfloor,\mathtt{I}^1_y[i] =\lfloor y_i^w/\mathtt{d}\rfloor,\mathtt{I}^1_z[i] =\lfloor z_i^w/\mathtt{d}\rfloor
\end{equation}
which is stored in a Structure of Array (SoA) layout $\mathtt{I}^1$ alongside their corresponding pose indices to ensure coalesced memory access. The points in $\mathtt{P}$ are then transformed into a point cluster representation and stored in $\mathtt{C}^1$, as defined by \Cref{eq:cluster}. Subsequently, we apply the standard primitive \texttt{sort\_by\_key} to sort the point clusters $\mathtt{C}^1$ by $\mathtt{I}^1$, grouping those associated with the same pose within the same voxel. This is followed by a \texttt{reduce\_by\_key} operation to aggregate the point clusters in $\mathtt{C}^1$ based on $\mathtt{I}^1$, significantly reducing memory requirements for storage. To enhance efficiency, data transfer and point clustering are performed asynchronously between the CPU and GPU using CUDA Streams. We designate this process as $\mathtt{L}_1$ voxelization, which serves as the foundation for a bottom-up voxelization approach that progresses to the $\mathtt{L}_k$ level.
\subsubsection{Bottom-up Voxelization}
\begin{algorithm}[t]
	\SetAlgoLined
	\SetKwInOut{Input}{Input}
	\SetKwInOut{Output}{Output}
	\SetKwInOut{Param}{Params}
        \SetKwFunction{multilevel}{$\mathtt{BottomUpVoxelization}$}
        \SetKwFunction{select}{$\mathtt{Selection}$}
        \SetKwFunction{map}{$\mathtt{Mapper}$}
        \SetKwFunction{voxelize}{$\mathtt{Voxelize}$}
	\SetKwFunction{cluster}{$\mathtt{ToCluster}$}
	\SetKwFunction{level}{$\mathtt{IndexUp}$}
	\SetKwFunction{concat}{$\mathtt{Concat}$}
        \SetKwFunction{transform}{$\mathtt{TransformToWorld}$}
        \SetKwFunction{voxelid}{$\mathtt{VoxelId}$}
        \SetKwFunction{pose}{$\mathtt{PoseId}$}
        \SetKwFunction{parselect}{$\mathtt{ParallelSelect}$}
        \SetKwFunction{const}{$\mathtt{ConstArray}$}
        \SetKwFunction{search}{$\mathtt{BinarySearch}$}
        \SetKwFunction{sort}{$\mathtt{sort\_by\_key}$}
        \SetKwFunction{reduce}{$\mathtt{reduce\_by\_key}$}
        \SetKwFunction{scan}{$\mathtt{exclusive\_scan}$}
        \SetKwFunction{scatter}{$\mathtt{scatter}$}
	\SetKwFunction{return}{$\mathtt{return}$}

	\SetKwProg{Fn}{Function}{}{}
        \SetKwProg{Par}{Parallel For}{}{}        
        
	\Param{Planar threshold $\tau$}
	\Input{Index $\mathtt{I^1}$, Cluster $\mathtt{C}^1$, Max Level $\mathtt{L}$\\}
    
	\Fn{\multilevel{$\mathtt{I^1}$,$\mathtt{C^1}$,$\mathtt{L}$}}{
         \For{$\mathtt{k}\gets 2$ $\mathtt{to}$ $\mathtt{L}$}{
            \Par{$\mathtt{I^{k-1}_i}$,$\mathtt{C^{k-1}_i}$ $\mathtt{in}$ $\mathtt{I^k}$,$\mathtt{C^k}$}{
                    $\mathtt{I^{k}_i}\gets$\level{$\mathtt{I^{k-1}_i}$} via Eq.~\eqref{eq:indexup}\\
                    $\mathtt{C^{k}_i}\gets \mathtt{C^{k-1}_i}$\\            
            }
            \sort{$\mathtt{I^k}$,$\mathtt{C^k}$}\\
            \reduce{$\mathtt{I^k}$,$\mathtt{C^k}$}\\
        }
        $\mathtt{cluster},\mathtt{idx}\gets$\concat{$\mathtt{C},\mathtt{I}$}\\
        \return{$\mathtt{cluster},\mathtt{idx}$}\\
    }
	\textbf{End Function}\\
    \Fn{\select{$\mathtt{cluster}$,$\mathtt{T}$}}{    $\mathtt{C}\gets$\transform{$\mathtt{cluster}$,$\mathtt{T}$}\\
        \reduce{\voxelid{$\mathtt{idx}$,$\mathtt{C}$}}\\
        \parselect{$\mathtt{C}$,$\tau$,$\mathtt{cluster}$}\\
        \return{$\mathtt{cluster}$}        
    }
	\textbf{End Function}\\
    \Fn{\map{$\mathtt{idx}$}}{
    $\mathtt{CNT}\gets$\const{$1$}\\
        \reduce{\voxelid{$\mathtt{idx}$},$\mathtt{CNT}$}\\
        $\mathtt{Id}\gets$\scan{$\mathtt{CNT}$}\\
        \Par{$\mathtt{i}$ $\mathtt{in}$ $\mathtt{idx}$}{
            $\mathtt{k}\gets$\search{$\mathtt{i}$, $\mathtt{Id}$}\\
            $\mathcal{M}_\mathtt{v}[i]\gets\mathtt{k}$,             $\mathtt{I_A},\mathtt{I_B}\gets \mathtt{i}$\\
        }
        $\mathtt{I_P}\gets$ \pose{$\mathtt{idx}$}\\
        \sort{$\mathtt{I_P}$,$\mathtt{I_A}$}\\        $\mathcal{M}_\mathtt{s}\gets$\scatter{$\mathtt{I_A}[i]$,$\mathtt{I_B}[i]$}\\
       \return{$\mathtt{I_P}$,$\mathcal{M}_\mathtt{v}$,$\mathcal{M}_\mathtt{s}$}
    }
	\textbf{End Function}\\    
	\caption{Preprocessing}
	\label{alg:preprocess}
\end{algorithm}
In contrast to conventional methods that employ an Octree structure\cite{liu2023efficient} to extract planar features through recursive top-down subdivision of point clouds, our approach begins at the minimal resolution $\mathtt{d}$ and directly merges clusters into higher-level voxels using their voxel indices. As the voxel resolution at the $k$-th level doubles that of the preceding level, the voxel indices for $\mathtt{L}_k$ voxelization are computed as follows:
\begin{equation}
\label{eq:indexup}
\begin{aligned}
        \mathtt{I}_{x/y/z}^k[i] = \lfloor \mathtt{I}_{x/y/z}^{k-1}[i]/2\rfloor\\
\end{aligned}
\end{equation}
Subsequently, we use \texttt{sort\_by\_key} and \texttt{reduce\_by\_key} operations to aggregate clusters at each level. The resulting clusters from different voxelization levels are concatenated into the SoA layouts \texttt{idx} and \texttt{clusters}. A pseudo code is given in \Cref{alg:preprocess}, Lines 1-11.

\subsubsection{Voxel Selection}
The voxel selection process, outlined in Lines 12–17, evaluates the planar features of each voxel by $\lambda_{0}/\lambda_1<\tau$ where $\lambda_0$ and $\lambda_1$ represent the smallest and second-smallest eigenvalues of the plane within a voxel, respectively, and $\tau$ denotes the planar threshold.

In the selection process, the point clusters stored in \texttt{cluster} are first transformed into the world coordinate system using \Cref{eq:cluster} (Line 13) and subsequently aggregated within the same voxel using the \texttt{reduce\_by\_key} primitive (Line 14). A lightweight parallel kernel, \texttt{ParallelSelect}, is then employed to identify and retain point clusters whose corresponding voxels exhibit planar features that satisfy the predefined threshold $\tau$. 

Notably, our bottom-up voxelization and selection strategy differs from traditional octree-based methods by capturing a greater number of planar features. Unlike top-down approaches, which may overlook planar features at higher-resolution voxels (lower levels) when lower-resolution voxels (higher levels) meet the threshold condition, our method incorporates features starting from the minimum resolution, achieving comprehensive feature extraction.

\subsubsection{Index Mapper}
Two index mappers $\mathcal{M}_\mathtt{v}$ and $\mathcal{M}_{\mathtt{s}}$ are prepared to facilitate the calculation of Jacobian and Hessian matrices as introduced in \Cref{subsubsec:calcJH}. $\mathcal{M}_\mathtt{v}$ returns the voxel index in which a given point cluster resides, while $\mathcal{M}_{\mathtt{s}}$ maps a cluster from a layout sorted by voxel index to a layout sorted by pose index. The algorithm is outlined in Lines 18-29. 

The mapper $\mathcal{M}_\mathtt{v}$ is constructed through the following procedure. First, we utilize \texttt{reduce\_by\_key}, using voxel indices as keys and a constant array of $1$ as values for counts. This operation outputs the number of clusters corresponding to each voxel (Lines 19-20). Next, we apply \texttt{exclusive\_scan} to determine the index of the first cluster within each voxel, denoted as \texttt{Id} (Line 21). Subsequently, we perform a parallel bisection search for each cluster to identify the corresponding voxel, yielding the results for $\mathcal{M}_V$ (Lines 23-24).

For the mapper $\mathcal{M}_\mathtt{s}$, we initially construct two increasing arrays, denoted as $\mathtt{I_A}$ and $\mathtt{I_B}$ (Line 24). Subsequently, $\mathtt{I_A}$ is sorted according to the pose indices (Lines 25–26). We then employ the $\texttt{scatter}$ operation to reorder $\mathtt{I_B}$, utilizing the previously sorted $\mathtt{I_A}$ as a mapping function (Line 27). The resulting output constitutes the mapper $\mathcal{M}_\mathtt{s}$.

\subsection{Optimization}
\label{subsec:opt}
\begin{algorithm}[t]
    \SetAlgoLined
    \SetKwInOut{Input}{Input}
    \SetKwInOut{Output}{Output}
    \SetKwInOut{Param}{Params}

    \SetKwFunction{sum}{$\mathtt{reduce}$}
    \SetKwFunction{calcres}{$\mathtt{CalRes}$}
    \SetKwFunction{res}{$\mathtt{Residual}$}
    \SetKwFunction{JH}{$\mathtt{CalJH}$}
    \SetKwFunction{calcJ}{$\mathtt{Jacobian}$}
    \SetKwFunction{calcH}{$\mathtt{Hessian}$}
    \SetKwFunction{solve}{$\mathtt{Solver}$}
    
    \SetKwFunction{opt}{$\mathtt{LM\_Optimizer}$}
    \SetKwFunction{eig}{$\mathtt{EigenVector}$}
    \SetKwFunction{diag}{$\mathtt{Diagonal}$}
    \SetKwFunction{ldlt}{$\mathtt{LDLT}$}
    \SetKwFunction{pred}{$\mathtt{PredResReduction}$}
    \SetKwFunction{adjust}{$\mathtt{Adjust}$}
    \SetKwProg{Fn}{Function}{}{}
    \SetKwProg{Par}{Parallel For}{}{} 
    \SetKwProg{alg}{Algorithm}{}{}
    \SetKwProg{while}{While}{}{}

\Input{Poses $\mathtt{T}$, Cluster $\mathtt{cluster}$, Indices $\mathtt{idx}$, Mapper $\mathcal{M}_V$, $\mathcal{M}_{\mathtt{sort}}$, Damping Ratio $\mu$}    

    \Fn{\calcres{$\mathtt{T}$,$\mathtt{cluster}$,$\mathtt{idx}$}}{
$\mathtt{C^W}\gets$\transform{$\mathtt{cluster}$,$\mathtt{T}$}\\  
        \reduce{\voxelid{$\mathtt{idx}$},$\mathtt{C^W}$}\\
        \Par{$\mathtt{C}_i$ $\mathtt{in}$ $\mathtt{C^w}$}{
            $\boldsymbol{u}[i]\gets$\eig{$\mathtt{C_i}$}\\
            $\mathtt{res}[i]\gets$\res{$\mathtt{C}_i$, $\boldsymbol{u}[i]$ }
        }
        $\mathtt{residual}\gets$\sum{$\mathtt{res}$}\\
        \return{$\mathtt{residual}$,$\boldsymbol{u}$}\\
    }
    \textbf{End Function}\\
    \Fn{\JH{$\mathtt{T}$,$\mathtt{cluster}$,$\mathtt{I_P}$,$\mathcal{M}_\mathtt{v}$,$\mathcal{M}_\mathtt{s}$,$\boldsymbol{\mu}$}}{
        \Par{$\mathtt{T}_i,\mathtt{C}_i$ $\mathtt{in}$ $\mathtt{T}$,$\mathtt{cluster}$}{
            $i_v\gets\mathcal{M}_\mathtt{v}[i]$,
            $i_p\gets\mathcal{M}_\mathtt{s}[i]$\\
            $\mathtt{J}[i_p]\gets$\calcJ{$\mathtt{T_i},\mathtt{C_i},\boldsymbol{u}[i_v]$}\\
            $\mathtt{H}[i_p]\gets$\calcH{$\mathtt{T_i},\mathtt{C_i},\boldsymbol{u}[i_v]$}\\
        }
        \reduce{$\mathtt{I_P}$,$\mathtt{J}$}\\
        \reduce{$\mathtt{I_P}$,$\mathtt{H}$}\\ 
        \return{$\mathtt{J}$,$\mathtt{H}$}
    }
    \textbf{End Function}\\
    \Fn{\solve{$\mathtt{T}$,$\mu$,$\mathtt{J}$,$\mathtt{H}$}}{
        \Par{$i\gets$ $1$ $\mathtt{to}$ $\mathtt{N_T}$}{
            $\boldsymbol{A}\gets$ $\mathtt{H}[i]+\mu$\diag{$\mathtt{H}[i]$}\\
            $\boldsymbol{b}\gets -\mathtt{J}[i]$,
            $\Delta \mathtt{T}\gets$\ldlt{$\boldsymbol{A}$,$\boldsymbol{b}$}\\
            $\mathtt{T}_i\gets\mathtt{T}_i+\Delta\mathtt{T}$
        }
        \return{$\mathtt{T}$}\\
    }
    \textbf{End Function}\\
            
    \caption{Optimization Process}
    \label{alg:opt}
    
\end{algorithm}
We describe the parallel computation to solve the optimization cost function in Eq.~\eqref{eq:surrogate}, with the pseudo-code presented in \Cref{alg:opt}.
\subsubsection{LM optimizer}
In our framework, the Levenberg–Marquardt (LM) algorithm serves as the optimizer, requiring the computation of residuals, Jacobian matrices, and Hessian matrices to solve for the increment and adjust the damping factor for optimization. 
In contrast to previous works that rely on sequential computation,
we compute these components in parallel as detailed in the following sections.
Furthermore, as the poses are decoupled through the majorization-minimization method, the increment solver can also be executed in parallel, 
further enhancing the efficiency of the optimization process.

\subsubsection{Residual Evaluation} The residual is computed through a parallel procedure similar to voxel selection, as detailed in Lines 1–9. Firstly, the clusters are transformed into the world coordinate using the pose from the current optimization step (Line 2). These are then aggregated via $\texttt{reduce\_by\_key}$ to derive the plane features for each voxel (Line 3), which are subsequently utilized to calculate the plane normal vector $\boldsymbol{u}$ (Line 5). The residual for each voxel is determined according to Eq.~\eqref{eq:surrogate} and summed in parallel through \texttt{reduce} (Line 7). Notably, the normal vector $\boldsymbol{u}$ is stored in global memory to facilitate the subsequent computation of the Jacobian and Hessian matrices.

\subsubsection{Jacobian and Hessian} 
\label{subsubsec:calcJH}
Owing to the efficient decoupling of majorization-minimization formulation, the computation of Jacobian and Hessian matrices can be isolated to each cluster and its associated pose, which is described in Lines 10–18. For each point cluster, the normal vector of its corresponding voxel is retrieved using the mapper $\mathcal{M}_\mathtt{v}$, and the Jacobian and Hessian matrices are calculated according to \cite{li2026efficient}. Utilizing the mapper $\mathcal{M}_\mathtt{s}$, the resulting matrices are organized into a layout sorted by pose index, which facilitates the subsequent $\texttt{reduce\_by\_key}$. This primitive efficiently sums the Jacobian and Hessian contributions from different clusters for each pose.

\subsubsection{Solver} The decoupled formulation allows for the parallel solution of the increment for each pose, as outlined in Lines 19–25. 

\section{Time Complexity Analysis}
\label{sec:analysis}
While our framework is built upon BALM3 in the theoretical aspect, the time complexity in parallel computation is completely different from that using sequential computation. According to Brent's theorem\cite{brent1974parallel}, the time complexity for parallel computing is determined by three factors: the total work in sequential $W(n)$, the depth in parallel computation $D(n)$, and the number of processing threads $N_\mathtt{Th}$. Here, $n$ is the number of data to be processed. The parallel time complexity $T_p$ is obtained as:
\begin{equation}
    T_p = \frac{W(n)}{N_\mathtt{Th}} + D(n)
\end{equation}

We provide the parallel time complexity for preprocessing, evaluation of residual, Jacobian, and Hessian matrices, and solver, respectively. 
\begin{theorem}
The total sequential work for preprocessing is $O(N_\mathtt{P}\log(N_\mathtt{P}))$ and the depth is $O(\log^2(N_\mathtt{P}))$. Therefore, the parallel time complexity for preprocessing is
\begin{equation}
    T_p^{\mathtt{pre}}=\frac{O(N_\mathtt{P}\log(N_\mathtt{P}))}{N_\mathtt{Th}}+O(\log^2(N_{\mathtt{P}}))
\end{equation}
where $N_{\mathtt{P}}$ is the number of raw points.
\end{theorem}

With $N_\mathtt{C}$ clusters to be processed, the parallel time complexity for computing residuals, Jacobian, and Hessian matrices is as follows:
\begin{theorem}
The total sequential work for residual, Jacobian and Hessian matrices computation is $O(N_\mathtt{C})$ and the depth is $O(\log(N_\mathtt{C}))$. Therefore, the parallel time complexity is
\begin{equation}
    T_p^{\mathtt{compute}}=\frac{O(N_\mathtt{C})}{N_\mathtt{Th}}+O(\log(N_{\mathtt{C}}))
\end{equation}
where $N_{\mathtt{C}}$ is the number of point clusters.
\end{theorem}

Finally, we provide the time complexity for the parallel solver of the increment in LM optimization as below:
\begin{theorem}
The total sequential work for parallel solver is $O(N_\mathtt{T})$ and the depth is $O(1)$. Therefore, the parallel time complexity for preprocessing is
\begin{equation}
    T_p^{\mathtt{solve}}=\frac{O(N_\mathtt{T})}{N_\mathtt{Th}}
\end{equation}
where $N_{\mathtt{T}}$ is the number of poses.
\end{theorem}

Due to space constraints, the proof of the parallel time complexity is provided in the supplementary materials$^1$. Based on the analysis above, the parallel time complexity is primarily governed by the first term, which depends on $N_\mathtt{Th}$, due to the limited number of threads available on modern GPUs. This finding is consistent with the runtime observations presented in Figure~\ref{fig:analysis}. Furthermore, the analysis also reveals that, even with unlimited computational resources, the time complexity remains lower-bounded by the second term corresponding to the depth of parallel computation.

\section{Benchmark Comparison}
 \label{sec:bench}
\begin{table*}[t]
\centering
\caption{Benchmark Comparison in Efficiency and Accuracy}
\setlength{\tabcolsep}{6.0pt}
\label{tab:benchmark}
\resizebox{\textwidth}{!}{%
\small
\begin{tabular}{@{}clcccclccccc@{}}
\toprule
\multirow{3}{*}{\centering Dataset} & \multicolumn{1}{c}{\multirow{3}{*}{Sequence}} & \multicolumn{5}{c}{Total Time (s)} & \multicolumn{5}{c}{RMSE (m)} 
\\ \cmidrule(l){3-7}  \cmidrule(l){8-12} 
 & \multicolumn{1}{c}{} & \multirow{2}{*}{\centering HBA} & \multirow{2}{*}{BALM3} & \multicolumn{2}{c}{Ours} & \multicolumn{1}{c}{\multirow{2}{*}{Speed$^2$$\uparrow$}} & \multirow{2}{*}{HBA} & \multirow{2}{*}{BALM3} & \multicolumn{2}{c}{Ours} & \multirow{2}{*}{$\Delta_{\mathtt{err}}$ $^3$$\downarrow$} \\
 & \multicolumn{1}{c}{} &  &  & \multicolumn{1}{l}{$\mathtt{L}=3$} & \multicolumn{1}{l}{$\mathtt{L}=1$$^*$} & \multicolumn{1}{c}{} &  &  & \multicolumn{1}{l}{$\mathtt{L}=3$} & \multicolumn{1}{l}{$\mathtt{L}=1$$^*$} &  \\ \midrule
\centering HeLiPR & Roundabout\_Avia01 & 214.63 & 152.04 & 22.15 & \textbf{17.53} & $_\times$6.86 & 2.085 & 1.722 & \textbf{1.583} & 1.968 & -0.1383 \\
 & Roundabout\_Avia02 & 197.99 & 61.42 & 10.74 & \textbf{1.56} & $_\times$5.72 & \textbf{1.384} & 1.471 & 1.444 & 2.473 & 0.0594 \\
 & Roundabout\_Avia03 & 175.75 & 185.43 & 18.59 & \textbf{12.92} & $_\times$9.97 & 2.099 & \textbf{1.870} & 2.010 & 2.306 & 0.1399 \\
 & Bridge\_Avia01 & 236.73 & 52.29 & 6.73 & \textbf{4.79} & $_\times$7.77 & \textbf{11.129} & 11.186 & 11.160 & 11.228 & 0.0311 \\
 & Bridge\_Avia02 & 210.72 & 19.78 & \textbf{2.52} & 3.11 & $_\times$7.84 & 6.576 & 6.440 & 6.445 & \textbf{6.432} & 0.0048 \\
 & Bridge\_Avia03 & 202.11 & 22.15 & 4.28 & \textbf{2.44} & $_\times$5.17 & \textbf{10.459} & 10.461 & 10.463 & 10.465 & 0.0040 \\ \midrule
\multirow{18}{*}{\centering MaRS-LVIG} & AMtown01 & 521.31 & 23.04 & \textbf{7.70} & 7.68 & $_\times$2.99 & 1.584 & 1.497 & \textbf{1.350} & 1.373 & -0.1463 \\
 & AMtown02 & 270.66 & 13.79 & \textbf{4.43} & 4.44 & $_\times$3.11 & 8.551 & 8.544 & \textbf{8.530} & 8.612 & -0.0142 \\
 & AMtown03 & 226.04 & 15.55 & \textbf{5.73} & 6.40 & $_\times$2.71 & 5.484 & 4.661 & \textbf{4.133} & 4.400 & -0.5285 \\
 & AMvalley01 & 469.57 & 19.52 & \textbf{11.86} & 14.26 & $_\times$1.65 & 6.530 & 6.099 & \textbf{6.062} & 6.067 & -0.0371 \\
 & AMvalley02 & 254.63 & 7.33 & 7.11 & \textbf{5.29} & $_\times$1.03 & 14.740 & \textbf{14.619} & 14.645 & 14.639 & 0.0257 \\
 & AMvalley03 & 178.19 & 15.30 & \textbf{2.20} & 2.38 & $_\times$6.97 & 17.725 & \textbf{17.684} & 17.689 & 17.698 & 0.0050 \\
 & HKairport01 & 313.63 & 7.72 & 4.80 & \textbf{4.19} & $_\times$1.61 & 0.640 & 0.649 & \textbf{0.555} & \textbf{0.555} & -0.0846 \\
 & HKairport02 & 159.41 & 5.19 & \textbf{1.51} & 1.83 & $_\times$3.44 & \textbf{0.544} & 0.610 & 0.575 & 0.575 & 0.0311 \\
 & HKairport03 & 136.17 & 1.72 & 1.18 & \textbf{0.85} & $_\times$1.46 & \textbf{0.698} & 0.815 & 0.710 & 0.727 & 0.0124 \\
 & HKairport\_GNSS01 & 299.25 & 5.55 & 2.25 & \textbf{1.81} & $_\times$2.47 & 0.444 & 0.462 & 0.442 & \textbf{0.441} & -0.0024 \\
 & HKairport\_GNSS02 & 159.69 & 4.50 & \textbf{1.24} & 1.46 & $_\times$3.63 & 0.637 & 0.682 & \textbf{0.629} & \textbf{0.629} & -0.0076 \\
 & HKairport\_GNSS03 & 136.51 & 5.00 & \textbf{2.13} & 2.32 & $_\times$2.35 & 0.539 & 0.548 & 0.534 & \textbf{0.529} & -0.0050 \\
 & HKisland01 & 145.40 & 2.95 & 1.51 & \textbf{1.36} & $_\times$1.95 & \textbf{0.556} & 0.593 & 0.560 & 0.560 & 0.0040 \\
 & HKisland02 & 83.60 & 2.16 & 1.30 & \textbf{1.16} & $_\times$1.66 & \textbf{0.707} & 0.734 & 0.709 & 0.708 & 0.0016 \\
 & HKisland03 & 80.28 & 2.68 & 1.27 & \textbf{1.22} & $_\times$2.11 & \textbf{0.843} & 0.952 & 0.897 & 0.897 & 0.0542 \\
 & HKisland\_GNSS01 & 166.57 & 2.71 & 1.57 & \textbf{1.35} & $_\times$1.72 & 0.619 & 0.682 & \textbf{0.599} & \textbf{0.599} & -0.0202 \\
 & HKisland\_GNSS02 & 94.67 & 2.52 & 1.43 & \textbf{1.31} & $_\times$1.76 & 0.741 & 0.814 & \textbf{0.728} & \textbf{0.728} & -0.0130 \\
 & HKisland\_GNSS03 & 92.97 & 3.01 & \textbf{0.65} & 0.74 & $_\times$4.60 & 1.085 & 1.107 & \textbf{1.074} & 1.060 & -0.0105 \\ \midrule
\multirow{9}{*}{\centering MulRan} & DCC01 & 301.73 & 10.07 & 1.66 & \textbf{0.94} & $_\times$6.08 & 5.391 & 5.354 & \textbf{5.260} & 5.470 & -0.0943 \\
 & DCC02 & 323.68 & 9.82 & 2.11 & \textbf{1.40} & $_\times$4.65 & 2.882 & 2.938 & \textbf{2.844} & 2.990 & -0.0382 \\
 & DCC03 & 417.16 & 12.48 & 2.55 & \textbf{1.89} & $_\times$4.90 & 2.114 & 2.113 & \textbf{2.011} & 2.201 & -0.1021 \\
 & KAIST01 & 313.43 & 10.03 & 2.18 & \textbf{1.58} & $_\times$4.61 & \textbf{3.292} & 3.351 & 3.323 & 3.352 & 0.0316 \\
 & KAIST02 & 333.03 & 9.91 & 2.38 & \textbf{1.90} & $_\times$4.17 & \textbf{3.353} & 3.401 & 3.389 & 3.409 & 0.0354 \\
 & KAIST03 & 447.44 & 10.95 & 2.38 & \textbf{1.82} & $_\times$4.60 & \textbf{2.905} & 2.930 & 2.924 & 2.939 & 0.0196 \\
 & Riverside01 & 238.47 & 4.88 & 1.28 & \textbf{0.79} & $_\times$3.80 & 8.461 & 8.425 & \textbf{8.410} & 8.420 & -0.0156 \\
 & Riverside02 & 252.76 & 7.60 & 1.20 & \textbf{0.72} & $_\times$6.32 & \textbf{11.410} & 11.428 & 11.443 & 11.447 & 0.0327 \\
 & Riverside03 & 345.43 & 11.32 & 2.69 & \textbf{1.89} & $_\times$4.22 & \textbf{12.984} & 13.082 & 13.089 & 13.082 & 0.1051 \\ \bottomrule
\end{tabular}%
}
\begin{tablenotes}
    \footnotesize
    \item $^*$ These two columns are only used for ablation study.
    \item $^1$ {Speed: Acceleration ratio of our method (denoted as ``Ours'') compared to the best-performing method between HBA and BALM3.}
    \item $^2$ {\(\Delta_{\mathtt{err}}\): The RMSE difference of our method relative to the best-performing method between HBA and BALM3.}
\end{tablenotes}
\end{table*}
In this section, we perform comprehensive benchmark comparisons in terms of efficiency and accuracy with state-of-the-art LiDAR bundle adjustment methods using public datasets, emphasizing large-scale data sequences.

\subsection{Experiment Setup}
\subsubsection{Dataset} The public datasets used in our experiments include \textit{HeLiPR}\cite{jung2024helipr}, \textit{MaRS-LVIG}\cite{li2024mars}, and \textit{MulRan}\cite{kim2020mulran}. These datasets encompass data collected from diverse environments and platforms, utilizing various types of LiDAR sensors. The specific data sequences for each dataset are detailed in the supplementary$^1$.

\subsubsection{Baselines}
 We compare our method with HBA~\cite{liu2023large}, a hierarchical LiDAR bundle adjustment framework designed for large-scale datasets, and BALM3~\cite{li2026efficient}, which provides the theoretical foundation of our method. 
 
\subsubsection{Computation Platform} The experiments of HBA and BALM3 are conducted on a high-performance computing desktop equipped with Intel-13900KF CPU and \SI{64}{\giga\byte} RAM.  Since our method requires substantial GPU memory, we utilized an A100 Tensor Core GPU with \SI{40}{\giga\byte} of on-board HBM2 VRAM.
\subsubsection{Parameters}
For HBA, we configure the minimal resolution to \SI{2.0}{\meter} with a total of 3 layers. Other parameters are adopted as defaults from its open-source code repository~\footnote{https://github.com/hku-mars/HBA}. For our method, the minimal voxel resolution is set to $\mathtt{d}$$=$\SI{4.0}{\meter} for the \textit{MulRan} and \textit{HeLiPR} datasets and to $\mathtt{d}$$=$\SI{3.0}{\meter} for the \textit{MaRS-LVIG} dataset. The voxelization level in our method is fixed at $\mathtt{L}=3$. For BALM3, we adopt the same minimum resolution as our method while it utilizes an adaptive voxelization strategy with a maximum of 3 layers.

\subsection{Efficiency Evaluation}
The evaluation results for efficiency and accuracy are presented in \Cref{tab:benchmark}. The total computation time includes both preprocessing and optimization phases for each method. Our method achieves the highest efficiency across all datasets, followed by BALM3, whereas HBA exhibits the lowest efficiency. 
Specifically, on the \textit{HeLiPR} dataset sequences, which contain over 20,000 poses and more than 300 million LiDAR points, BALM3 requires up to \SI{185.43}{\second} while our method takes only \SI{22.15}{\second}, i.e. about ten times faster. 
The other baseline, HBA demonstrates significantly longer computation times than BALM3, up to \SI{236.73}{\second} for the \textit{Bridge Avia01} sequence. 
Meanwhile, on the \textit{MulRan} dataset with over 200 million LiDAR points and 10,000 poses, our approach processes sequences in up to \SI{2.69}{\second}, compared to BALM3’s up to \SI{12.48}{\second} (around 4.6 time slower) and HBA’s up to \SI{447.44}{\second} (around 166 times slower). 
For the \textit{MaRS-LVIG} dataset, characterized by a significantly smaller number of LiDAR points and poses, our method achieves the greatest performance, slightly outperforming BALM3 and surpassing HBA by a substantial margin.

We also report the acceleration ratio (i.e., speed in \Cref{tab:benchmark}), computed as: $\mathtt{Speed} = \min({\mathtt{t}_{\mathtt{BALM3}},\mathtt{t}_{\mathtt{HBA}}})/\mathtt{t}_\mathtt{ours}$.
The results indicate that our method achieves a maximum acceleration of 9.97 times that of BALM3 for \textit{Roundabout\_Avia03} sequence.

\subsection{Accuracy Evaluation}
In terms of accuracy, we evaluate the root mean square error (RMSE) of the absolute position error (APE) using \textit{evo}\footnote{https://github.com/MichaelGrupp/evo} for assessment. Our method achieves the lowest RMSE for 17 out of 33 sequences tested, followed by HBA, which performs best in 13 sequences, and BALM3, which leads in 3 sequences. The modest accuracy improvements of our method primarily result from the bottom-up voxelization design. Owing to the efficient parallel processing framework, a larger number of plane features across varying voxel resolutions are incorporated into the optimization cost function, thereby applying additional constraints. In contrast, BALM3 and HBA utilize an adaptive voxelization approach, potentially considering fewer plane features. However, it should be noted that incorporating more planar features into the optimization cost function does not always enhance the final results, as demonstrated by our ablation study in \Cref{subsec:ablate}.

 \subsection{Time Analysis}
\begin{figure}
    \centering
\setlength\abovecaptionskip{-0.3\baselineskip}    
    \includegraphics[width=0.9\linewidth]{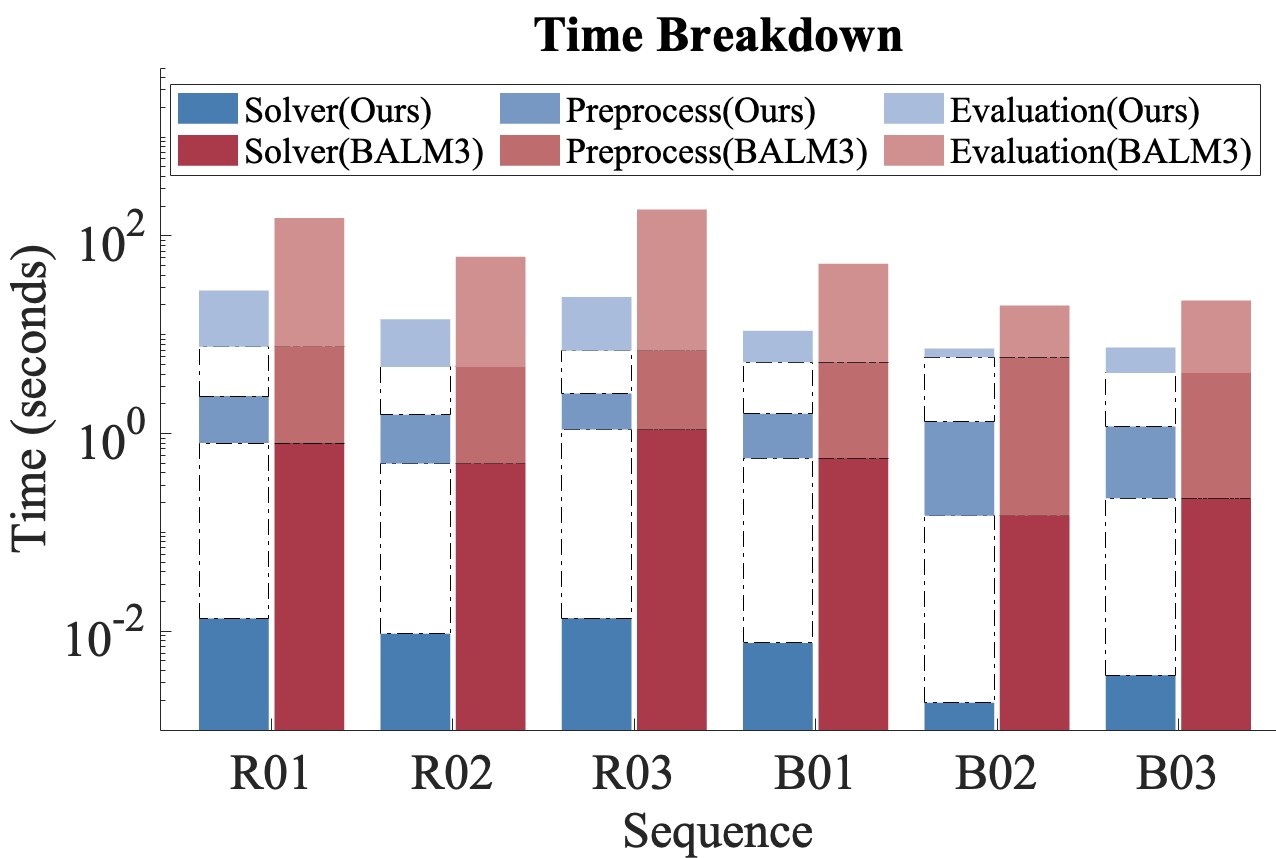}
    \caption{
    Time Breakdown Comparison of Our Method and BALM3 for the \textit{HeLiPR} Dataset. The bars represent preprocessing, evaluation, and solver times for six sequences (R01 to R03 and B01 to B03, corresponding to \textit{Roundabout\_Avia01} to \textit{Roundabout\_Avia03} and \textit{Bridge\_Avia01} to \textit{Bridge\_Avia03}, respectively). Blue bars for our method are shifted to align with the red bars for BALM3 for a clearer visual comparison of their values.}
    \label{fig:breakdown}
\end{figure}
We first present a detailed time breakdown of our method and BALM3, encompassing preprocessing time, evaluation time, and solver time for six sequences from the \textit{HeLiPR} dataset, as illustrated in \Cref{fig:breakdown}. The solver time for computing the increment in our method is notably small, averaging \SI{0.008}{\second}, compared to BALM3, which averages \SI{0.554}{\second}, resulting in an acceleration of approximately $70\mathtt x$. For preprocessing, our method requires an average of \SI{1.2}{\second}, while BALM3 requires \SI{5.18}{\second}, achieving an acceleration of $4.32$ times. Evaluation constitutes the most time-intensive phase of the process which includes the computation of residuals, Jacobian matrices, and Hessian matrices. Our method averages \SI{9.61}{\second}, outperforming BALM3, which averages \SI{76.45}{\second}, with an acceleration of roughly $8$ times.

    \begin{figure}
        \centering
    \setlength\abovecaptionskip{-0.5\baselineskip}    
        \includegraphics[width=1.0\linewidth]{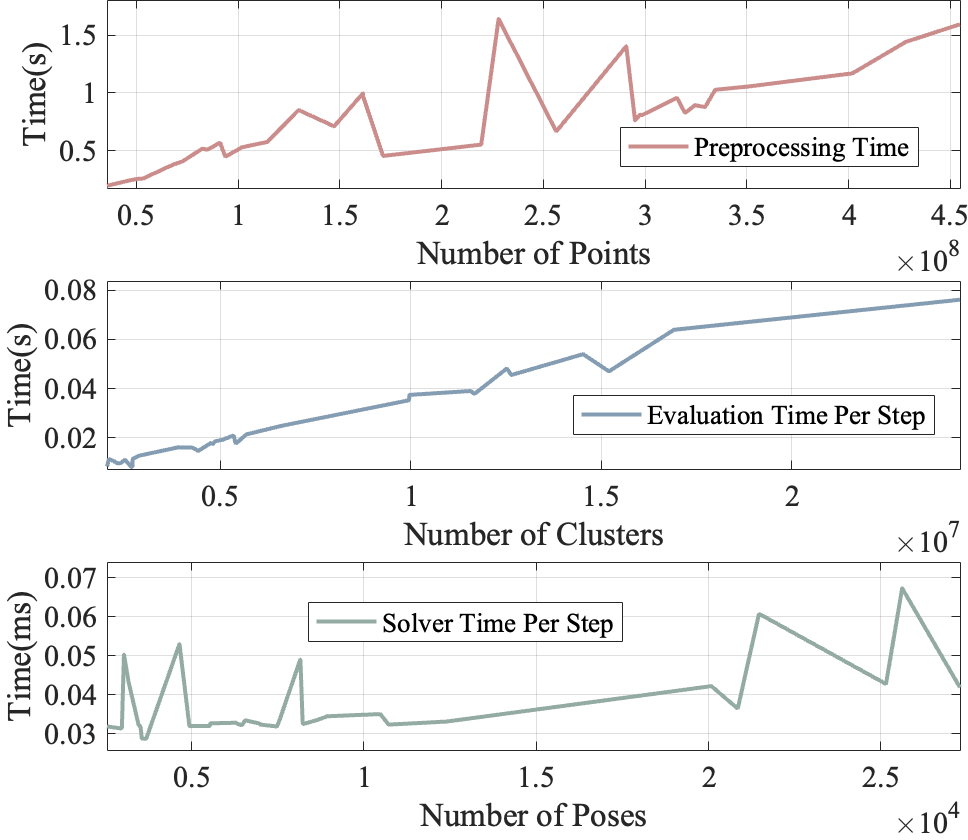}
        \caption{Experimental Time Analysis
    }
        \label{fig:analysis}
        
    \end{figure}

To validate the time complexity analysis presented in Section~\ref{sec:analysis}, we evaluated the preprocessing time, per-step evaluation time, and per-step solver time as functions of the number of points, clusters, and poses, respectively. As shown in Figure~\ref{fig:analysis}, the runtime exhibits a linear increase, indicating that the time complexity is primarily governed by sequential work, aligning well with our analysis. This is due to the limited number of parallel threads relative to the large volume of data. These results suggest that further improvements in time efficiency could be achieved by leveraging additional GPUs to enhance parallel computation.
\subsection{Ablation Study on Voxelization Levels}
\label{subsec:ablate}
Since our method employs a bottom-up voxelization with multilevel, we conduct a straightforward ablation study to compare the performance of multiple levels against a single level (i.e., $\mathtt L=1$) within our framework. As shown in \Cref{tab:benchmark}, incorporating additional voxels at varying resolutions enhances accuracy, albeit at the expense of increased processing time. However, we observed slight performance declines (e.g., \textit{HKairpot\_GNSS01}) and instances where performance did not improve (e.g., \textit{HKairpot01} and \textit{HKairpot\_GNSS03}). We emphasize that utilizing more voxels at lower resolutions requires careful consideration of the planar feature threshold to prevent including suboptimal planar features that could undermine the optimization results.


\section{Conclusion and Future Work}
This paper introduces the first parallel computing framework for LiDAR bundle adjustment. Within our framework, an adaptive asynchronous loading strategy mitigates GPU memory constraints in processing large-scale datasets, followed by a fully parallelized preprocessing method for planar features extraction. Subsequently, it leverages the majorization-minimization formulation to enable parallel optimization, alongside a parallel solver for the increment. The experiment results demonstrate substantial improvements in efficiency while maintaining accuracy. Future work will extend the framework to multi-GPU architectures to further scale global optimization and maximize efficiency.


\bibliographystyle{IEEEtran}
\bibliography{reference}

\end{document}